\documentclass[letterpaper]{article} 
\usepackage[margin=1in]{geometry}

\usepackage{times}  
\usepackage{helvet}  
\usepackage{courier}  
\usepackage[hyphens]{url}  
\usepackage{graphicx} 
\usepackage{natbib}  
\usepackage{caption} 
\usepackage{amssymb} 
\usepackage{booktabs} 
\newcommand{\Acal}{\mathcal{A}}
\newcommand{\E}{\mathbb{E}}

\newcommand{\Lcal}{\mathcal{L}}
\newcommand{\Dcal}{\mathcal{D}}
\usepackage{amsmath}

\newcommand{\R}{\mathbb{R}}
\usepackage{cleveref}

\usepackage{algorithm}
\newcommand{\Sbf}{\mathbf{S}}
\usepackage{algorithmicx}
\usepackage{algcompatible}
\usepackage{amsmath,graphicx,tikz}
\usetikzlibrary{positioning}
\usetikzlibrary{arrows.meta}
\usepackage{algorithm,algpseudocode}   
\algrenewcommand\algorithmiccomment[1]{\hfill{\small$\triangleright$~#1}} 
\algrenewcommand\textproc{} 
\usepackage{newfloat}
\usepackage{listings}
\DeclareCaptionStyle{ruled}{labelfont=normalfont,labelsep=colon,strut=off} 
\floatstyle{ruled}
\newfloat{listing}{tb}{lst}{}
\floatname{listing}{Listing}
\usepackage{authblk} 
\title{Ensuring Safety in Automated Mechanical Ventilation through Offline Reinforcement Learning and Digital Twin Verification}

\author[1]{Hang Yu\thanks{Hang Yu and Huidong Liu contributes equally to this work}}
\author[2]{Huidong Liu\textsuperscript{*}}
\author[3]{Qingchen Zhang$^\dagger$}
\author[1]{William Joy}
\author[4]{Kateryna Nikulina}
\author[4]{Andreas A. Schuppert}
\author[1]{Sina Saffaran$^\dagger$}
\author[1]{Declan Bates\thanks{Corresponding author: Sina Saffaran (Email: Sina.Saffaran.1@warwick.ac.uk), Declan Bates (D.Bates@warwick.ac.uk)}}

\affil[1]{School of Engineering, University of Warwick, Coventry, UK}
\affil[2]{School of Computer Science, Chongqing University, China}
\affil[3]{School of Computer Science and Technology, Hainan University, China}
\affil[4]{Institute for Computational Biomedicine, RWTH Aachen University, Germany}

\date{}  

\begin{document}

\maketitle
\pagestyle{plain}
\begin{abstract}
Mechanical ventilation (MV) is a life-saving intervention for patients with acute respiratory failure (ARF) in the ICU. However, inappropriate ventilator settings could cause ventilator-induced lung injury (VILI). Also, clinicians workload is shown to be directly linked to patient outcomes. Hence, MV should be personalized and automated to improve patient outcomes.
Previous attempts to incorporate personalization and automation in MV include traditional supervised learning and offline reinforcement learning (RL) approaches, which often neglect temporal dependencies and rely excessively on mortality-based rewards. As a result, early stage physiological deterioration and the risk of VILI are not adequately captured. To address these limitations, we propose Transformer-based Conservative Q-Learning (T-CQL), a novel offline RL framework that integrates a Transformer encoder for effective temporal modeling of patient dynamics, conservative adaptive regularization based on uncertainty quantification to ensure safety, and consistency regularization for robust decision-making. We build a clinically informed reward function that incorporates indicators of VILI and a score for severity of patients illness.
Also, previous work predominantly uses Fitted Q-Evaluation (FQE) for RL policy evaluation on static offline data, which is less responsive to dynamic environmental changes and susceptible to distribution shifts. To overcome these evaluation limitations, interactive digital twins of ARF patients were used for online “at the bedside” evaluation.
Our results demonstrate that T-CQL consistently outperforms existing state-of-the-art offline RL methodologies, providing safer and more effective ventilatory adjustments. Our framework demonstrates the potential of Transformer-based models combined with conservative RL strategies as a decision support tool in critical care.

\end{abstract}

\ifdefined\aaaianonymous
\else

\fi

\ifdefined\aaaianonymous

\fi
\section{Introduction}
Mechanical ventilation (MV) is one of the most important life-saving interventions for patients with acute respiratory failure (ARF) in the intensive care unit (ICU). However, MV itself can cause or exacerbate lung injury by applying excessive airway pressures and volumes, a phenomenon collectively known as ventilator-induced lung injury (VILI). Despite its importance, MV treatment strategies are largely based on general protocols derived from clinical trials. In practice, ventilator settings are manually adjusted by highly time-pressured clinicians \citep{r1}. Personalized and lung-protective treatment requires timely interventions in response to changes in the patient’s physiological state, which cannot be achieved within the constraints of current ICU workflows. A recent study found that important oxygen saturation targets were only achieved 40\% of the time \citep{r2, r3}. Another study also reported low adherence to current lung-protective ventilation guidelines, with only about 37–42\% of patients receiving lung-protective ventilation \citep{r38}.

 Lung-protective ventilation aims to minimize VILI \citep{r4} which, along with ensuring adequate gas exchange, is recognized as the primary therapeutic goals of MV. Current VILI minimization targets are usually focused on lower tidal volumes (typically 6 ml/kg predicted body weight) and limiting plateau pressure to $<$ 30~cmH$_2$O \citep{r5}. Pressure-controlled ventilation (PCV) is directly aligned with the primary therapeutic goal of minimizing VILI by allowing direct control of airway pressures, thus reducing the risk of excessive alveolar stretch \citep{r6}. This approach offers a more lung-protective strategy in situations where limiting peak and plateau pressures is prioritized \citep{r7}.

Previous work has attempted to demonstrate the potential of using machine learning (ML) to optimize ventilation treatments. These approaches have employed deep supervised learning \citep{r8} or tabular reinforcement learning (RL) \citep{r9}, either by mimicking the clinician's policy or utilizing fixed state spaces through clustering methods. However, these methods may result in suboptimal treatment outcomes, due to the ignorance of the sequential nature of ventilation and their inability to effectively represent continuous state spaces. Existing deep RL methods for optimizing MV parameters have primarily employed conservative offline RL \citep{r10, r11} to optimize MV parameters to ensure safety by avoiding overestimation of out-of-distribution states or actions.

However, these approaches \cite{r10, r11, r15, r16, r17} have several limitations: 1) They typically do not capture temporal characteristics, such as fluctuations in oxygenation, respiratory rate (RR), and heart rate (HR) over time, which are critical to detecting deterioration in the patient. Recent studies \citep{r12, r13} have demonstrated that incorporation of this dynamic state information, or a summary of the patient’s recent clinical history, can significantly improve the robustness of policy in critical care medicine. 2) These methods relied primarily on mortality-based rewards, either sparse or shaped. However, medical studies \citep{r14, r7, r18, r20} indicate that mortality is not a reliable indicator of the quality of MV treatment and cannot capture the subclinical or early stage effects of VILI, which can cause long-term complications even in survivors. 3) All previous works, to the best of our knowledge, have attempted to use Fitted Q-Evaluation (FQE) for evaluation of RL policy based on offline static data, which is less responsive to changes in the environment and may suffer from distribution shift. A recent survey \citep{r19} on RL in critical care medicine confirmed that all RL-based technologies for critically ill patients rely on retrospective off-policy evaluation (OPE), creating significant obstacles for the clinical implementation of RL-based mechanical ventilation optimization methods.

To address the aforementioned limitations in RL based automatic mechanic ventilation, we proposed a safety-guaranteed Transformer-based Conservative Q-Learning (T-CQL) for dynamic treatment regime validated through digital twin simulation. The key contributions of this work include:
\begin{itemize}
\item \textbf{Comprehensive Modeling of Patient Dynamic Characteristics.} We proposed T-CQL framework, which integrates Transformer architecture with conservative Q-Learning to address the limitations of conventional offline RL in dynamic treatment regimes by incorporating historical data representation with current state information. The architecture consists of three main components: a Transformer encoder for temporal feature extraction, a multi-head Q-network for action-value estimation, and uncertainty quantification modules for safe policy learning.
\item \textbf{Early Detection of the Risk of VILI.} We proposed a clinically relevant reward function that integrates indicators related to VILI and mortality prediction, such as driving pressure and Apache-II score, as intermediate rewards. This enables the model to effectively capture the subclinical or early-stage effects of VILI in PCV.
\item \textbf{Validation Using ``At the Bedside" Platform.} To validate offline RL in a clinical setting, we used a digital twin model of individual patients based on a high-fidelity computational model of the cardiopulmonary system, simulating physiological responses to ventilator settings. Using 98 ``digital twins," we demonstrate the effectiveness of our proposed methods in realistic clinical settings, which were benchmarked against existing approaches.
\end{itemize}

\section{Background \& Problem Formulation}
\subsection{Problem Definition}
Personalized and automated MV treatment can be formulated as a Markov Decision Process (MDP) in an offline RL setting. At each time step (e.g. every few hours of ICU ventilation), the patient’s state $s_t$ encapsulates their current clinical status (vitals, lab results, etc.), and the agent selects a ventilator action $a_t$ (adjustments to ventilator settings). The goal is to learn a policy $\pi(a_t \mid s_t)$ that maximizes cumulative reward, which is designed to correlate with a ``safe ventilation strategy" to minimize the risk of VILI by avoiding applying excessive pressures, volumes and energy to the lungs \citep{r5, r6}. Thus, the RL agent must behave conservatively, avoiding drastic or out-of-distribution actions that could jeopardize patient safety. We specifically target clinical objectives such as minimizing mortality (e.g. maximizing 90-day survival, reducing morality prediction score), while maintaining lung-protective ventilation using low driving pressure to reduce the risk of VILI \citep{r7}. By formulating the ventilation adjustment as an offline RL problem, the agent can leverage retrospective clinical data to optimize patient outcomes, thus avoiding active experiments on patients, which is ethically impermissible.
\subsection{State and Action Space Definitions}
State Space: Each state $s_t \in \mathcal{S}$ is a high-dimensional vector summarizing the patient’s current condition. We derive state features from a selection of vital signs, laboratory measurements, and other clinical observations recorded in the ICU databases. In total, we consider on the order of a few indicative variables that domain experts deem relevant to ventilator management and patient status. The state representation includes: 

\textbf{Demographics \& History}: Patient age, sex, weight, and comorbidity scores. 

\textbf{Vital Signs \& Scores}: Key hemodynamic and respiratory indicators such as heart rate, blood pressure (systolic, diastolic and mean), body temperature, oxygen saturation (SpO$_2$), and respiratory rate.

\textbf{Laboratory Values}: Frequent blood gas and metabolic lab results that reflect pulmonary and metabolic status, e.g. arterial blood pH, PaO$_2$, PaCO$_2$ (partial pressures of O$_2$ and CO$_2$), serum lactate (if available), electrolytes (Na, K, Cl, bicarbonate), renal function markers (creatinine, BUN), complete blood count (hemoglobin, WBC, platelet). These lab results provide insight into the patient’s current physiological derangements.

We do not explicitly include the ventilator settings in the state, since those are the decision variables the agent will control (their influence is captured via the patient’s physiological response). However, the state does reflect any downstream effects of previous ventilator settings (e.g. PaO$_2$, PaCO$_2$, etc.). This comprehensive state representation (derived from MIMIC data elements) equips the RL agent with essential information to evaluate patient status and clinical needs at time step $t$. 
The set of actions at each decision step is a set of ventilator settings. We define a discrete action space consisting of combinations of key ventilator parameters, focusing on those that have the most impact on gas exchange and lung mechanics. In our formulation, an action $a_t \in \mathcal{A}$ is a 5-dimensional tuple:
\[
\begin{aligned}
    &\quad\textbf{PEEP}:\;\text{Positive end‐expiratory pressure (cmH\textsubscript{2}O)},\\
    &\quad\textbf{FiO\textsubscript{2}}:\;\text{Fraction of inspired oxygen (\%)},\\
    &\quad\textbf{RR}:\;\text{Respiratory rate (breaths/min)},\\
    &\quad\textbf{I \!:\! E ratio}:\;\text{Inspiratory–expiratory time ratio},\\
    &\quad\textbf{Pvent}:\;\text{Pressure support level (cmH\textsubscript{2}O)}.
\end{aligned}
\]

These settings directly influence the delivered tidal volume, airway pressures, and mechanical power. Each of these five settings is discretized into a finite set of values covering the clinically reasonable range for patients with ARF. This resulted in a multi-dimensional action space of 13,440 discrete actions $(6 \times 8 \times 7 \times 5 \times 8) $. Driving pressure (DP) is defined as the difference between the plateau pressure $(P_{\mathrm{plat}}$ and PEEP in volume‐controlled ventilation. In PCV, where \(P_{\mathrm{plat}}\) is not directly measured, the driving pressure is approximated as the difference between the Peak Inspiratory Pressure (PIP) and PEEP, i.e., $\mathrm{DP} \approx \mathrm{PIP} - \mathrm{PEEP}$, which is a key indicator of VILI.

\subsection{Offline Reinforcement Learning}
RL can be classified into online and offline paradigms. Unlike online RL which involves an agent actively interacting with the environment by continuously updating policies from new experiences, offline RL learns from a precollected static dataset of transitions $D = \left\{ (s_t^i, a_t^i, r_t^i, s_{t+1}^i) \right\}_{i=1}^N$ without actively interacting with the environment \citep{r30, r31}. This makes offline RL particularly suitable for healthcare settings, where the risks and ethical concerns of real-time policy adjustments through direct patient interaction make online RL approaches impractical. When using offline RL in critical care, the policies made by the model solely depend on previously collected medical records. This leads offline RL to exhibit overconfidence in rarely seen or unseen scenarios \citep{r30}. Such overconfidence poses serious risks, with the agent potentially selecting high Q values learned from sparse data that may not generalize well and could endanger patients. For example, offline RL could suggest aggressive ventilatory settings (such as excessively high inspiratory pressures) merely because these interventions were associated with improved 90-day mortality within narrow and biased historical data, which could significantly increase the risk of VILI and causing long-term complications even in survivors.

\begin{figure}[htbp]
  \centering
  \includegraphics[width=0.47\textwidth]{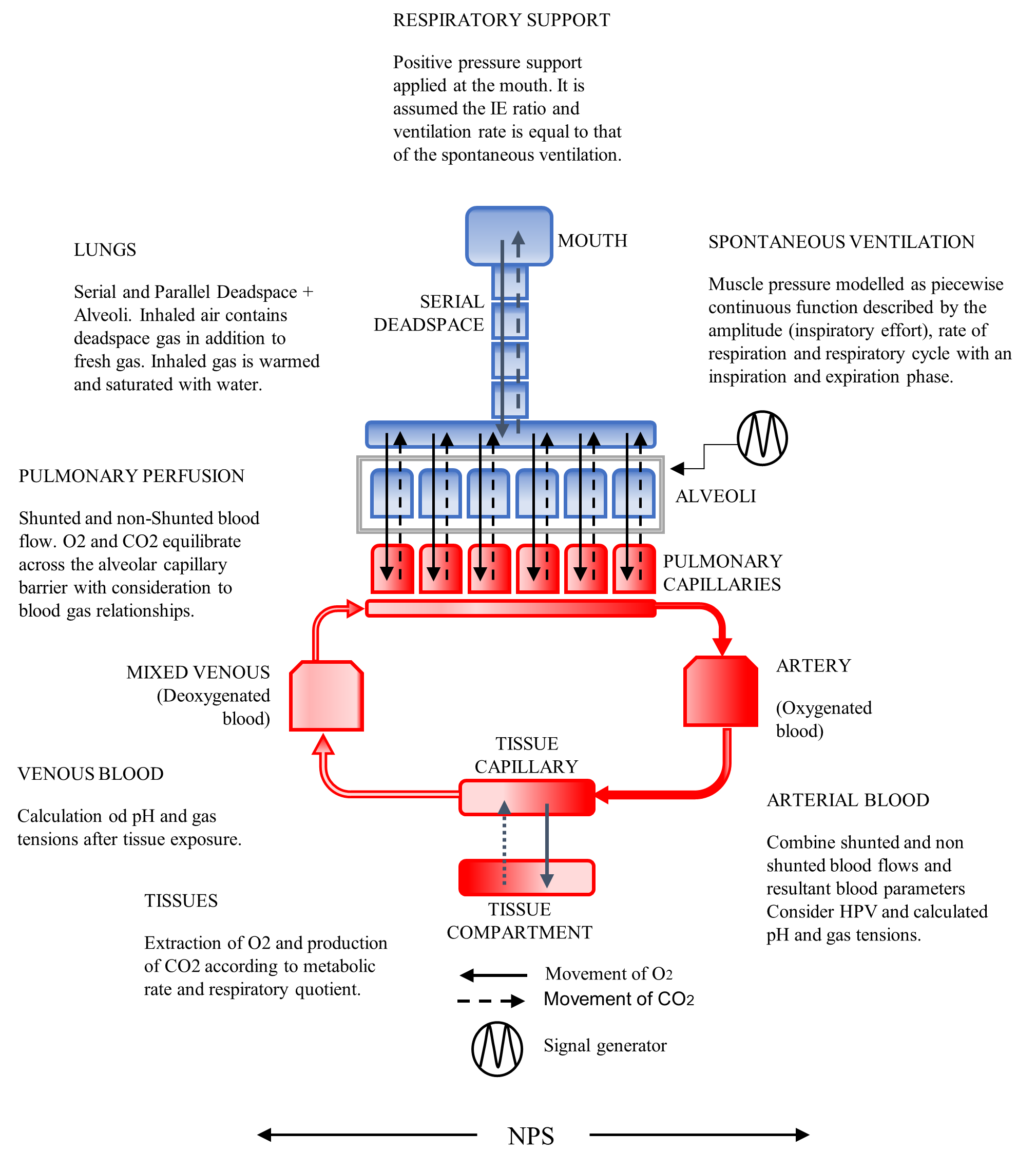}
  \caption{A simplified, diagrammatic representation of the digital twin cardiopulmonary system.}
  \label{fig:pulmonary_model}
\end{figure}

\subsection{Mechanistic Digital Twins}
The cardiopulmonary simulator used in this study to emulate real patient responses has been in continuous development over the past several years \citep{r40, r41, r42, r23, r25, r26, r27}, and has been used and validated in a number of different studies \citep{r22, r39, r28, r29}. It is based on a high-fidelity computational representation of the cardiopulmonary system, which simulates physiological responses to various ventilator settings. The model is organized as a system of several components, each component representing different sections of pulmonary dynamics and blood gas transport, e.g. the transport of air in the mouth, the tidal flow in the airways, the gas exchange in the alveolar compartments and their corresponding capillary compartment, the flow of blood in the arteries, the veins, the cardiovascular compartment, and the gas exchange process in the peripheral tissue compartments. Each component is described as several mass conserving equations, solved in series in an iterative manner. At the end of the iteration, the results of the solution of the final equations determine the independent variables of the first equation for the next iteration.
Figure~\ref{fig:pulmonary_model} shows a simplified diagrammatic representation of the model components and their interactions.

We generated 98 digital twins of patients with ARF and receiving mechanical ventilation by drawing distinct initial conditions (lung compliance, airway resistance, baseline blood‐gas values, etc.) from a parameter space validated and calibrated with computational simulator based on real patient data from Aachen University Hospital.

The system dynamics of the digital twins are modeled as:
\begin{equation}\tag{1}
s_{t+1} = f(s_t, a_t) + \omega_t,
\end{equation}
where the nonlinear state transition function \(f: \mathcal S \times \mathcal A \to \mathcal S\) defines the system’s deterministic evolution, and \(\omega_t\) represents process noise accounting for model‐to‐reality variations. The state transition probability \(P(s_{t+1}\mid s_t,a_t)\) depends on the distribution of \(\omega_t\) centered at \(f(s_t,a_t)\). While the mean evolution follows \(f(s_t,a_t)\), the stochastic term \(\omega_t\) enables the RL agent to develop robust control strategies under physiological uncertainty.  

\section{Method}
\begin{figure*}[!htbp]
  \centering
  \includegraphics[width=1.02\textwidth]{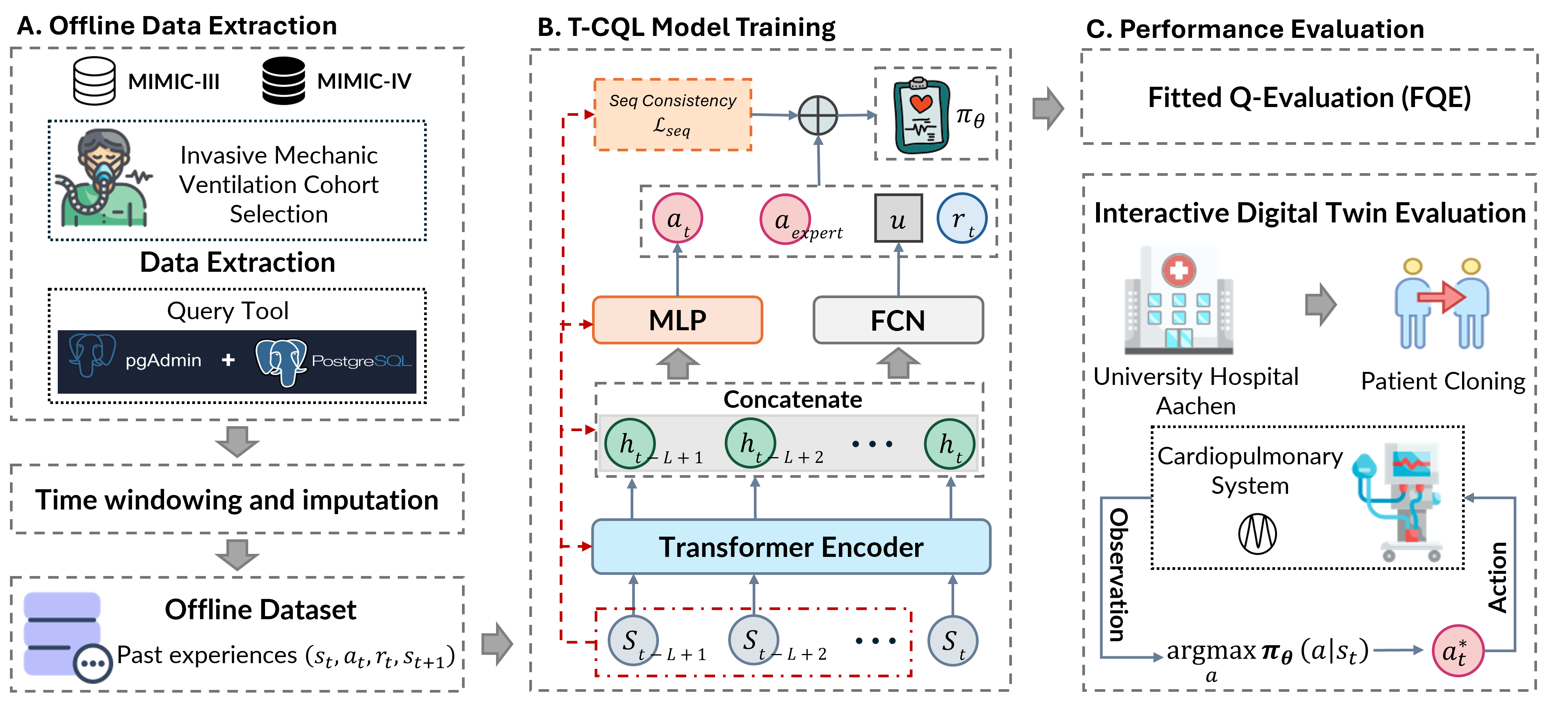}
  \caption{Overview of the proposed method. (A) Preparation of the structured offline dataset. (B) The training procedure of our proposed T-CQL method, which integrates a Transformer encoder for effective temporal modeling of patient dynamics, state-dependent conservative adaptive regularization based on uncertainty quantification to ensure safety, and consistency regularization for robust decision-making. (C) Evaluation methods: off-policy FQE and online interactive digital twin evaluation.}
  \label{fig:Overall architecture}
\end{figure*}
\subsection{Transformer‐based Conservative Q-Learning}
\subsubsection{Transformer‐based Q‐Network Encoder}
We parameterize the action‐value function $Q_\theta$ with a Transformer encoder (\texttt{TransformerDQN}) over a history of length $L$.  Given a historic states $\Sbf^{1:L}_t = (s_{t-L+1},\dots,s_t)\in\R^{L\times d}$, we can compute the hidden state representation:   
\begin{align}
  H &= \mathrm{TransformerEncoder}\bigl(\Sbf^{1:L}_t\bigr)\in\R^{L\times h}.
\end{align}
We build a single feature vector $\phi = [z_t \, \| \, \bar{z}] \in \mathbb{R}^{2h}$ by concatenating current‐step embedding $z_t = H_{t}\in\R^{h}$ (the last row of $H$ corresponding to the most recent state $S_t$) with history summary embedding $\bar z = \frac1L\sum_{i=1}^L H_i \in R^h$. For Q value prediction, we feed the feature vector $\phi$ into a feed-forward multilayer perception (MLP) ($\mathbb{R}^{2h} \rightarrow \mathbb{R}^{|\mathcal{A}|}$):
\begin{align}
Q_\theta(\mathbf{s}_t^{1:L},a) = \bigl[\mathrm{MLP}(\phi)\bigr]_a, a\in\mathcal{A},
\end{align}
where $\mathcal{A}$ denotes the MLP output logits in the action space, and $\theta$ denotes all trainable weights in the embedding layers, Transformer and MLP. To quantify epistemic uncertainty, we add a parallel ``uncertainty head" $\psi_{\theta}:\mathbb{R}^{h} \rightarrow \mathbb{R}$ and define:
\begin{align}
  u_\theta(\Sbf^{1:L}_t) \;=\;\mathrm{Var}\bigl(\{\psi_\theta(H_i)\}_{i=1}^L\bigr),\
\end{align}
where $\psi_{\theta} : \mathbb{R}^{h} \rightarrow \mathbb{R}$ is a learned linear head. High variance means the model is “unsure” across time-steps. Therefore, the states with higher estimated uncertainty receive exponentially stronger penalties for unseen actions. After a forward pass through \texttt{TransformerDQN}, we obtain both $Q_{\theta}(\mathbf{S}_{t_{1}:L}, a)$ for every action $a$, and the scalar $u_{\theta}(\mathbf{S}_{t_{1}:L}) \in \mathbb{R}$, which will later be used for computing a per-state weight as an adaptive T-CQL coefficient.

\subsubsection{T-CQL with Adaptive Regularization}
Building on CQL \cite{r31}, we combine a Bellman‐error term with an uncertainty‐aware conservative penalty. We define the temporal‐difference loss (mean squared Bellman error) as follows:
\begin{align}
  \Lcal_{\rm TD}
  = \E_{(s,a,r,s')\sim\Dcal}
    \bigl[Q_\theta(\Sbf^{1:L},a) - \bigl(r + \gamma\,Q_{\bar\theta}(\Sbf'^{1:L},a')\bigr)\bigr]^2,
\end{align}
where $a'=\arg\max_{a'}Q_\theta(\Sbf'^{1:L},a')$, $\gamma$ is a discount factor, and $\bar\theta$ is a target network. Then, for each state sequence, we define:
\begin{align}
  \ell_{\text{t-cql}}(\Sbf) = \tau \log 
\underbrace{ \sum_{a} \exp\left(Q_{\theta}(\Sbf^{1:L}, a) / \tau \right) }_{\text{soft-max over actions}}
- Q_{\theta}(\Sbf^{1:L}, a_{\text{data}}),
\end{align}
where $a_{\rm data}$ is the recorded action (the clinician’s chosen ventilator setting at state $S_i$) and $\tau>0$ is a temperature hyperparameter which controls the sharpness of the soft-max (appearing once inside the exponent as a divisor and once outside as a multiplier of the logarithm). The first term in the T-CQL regularizer can be considered as a soft-max approximation of $\max_{a}Q_\theta(s,a)$. Unlike the original CQL, we introduce an adaptive weighting coefficient to weight the penalty regularizer:
\begin{align}
   \Lcal_{\rm T-CQL}
  = \E_{s\sim\Dcal}\bigl[\underbrace{\alpha_0 \,\exp\bigl(\beta\,u_\theta(\Sbf^{1:L})\bigr)}_{\text{adaptive coefficient}}\,\ell_{\rm t-cql}(\Sbf)\bigr],
\end{align}
where the adaptive coefficient is used to quantify epistemic uncertainty. In contrast to the original CQL's regularizer that uniformly penalizes all historic states for the rare actions in the dataset, often leading to overly conservative policies, our adaptive regularizer introduces a state-dependent penalty. It acts as a state-dependent safety filter by penalizing unfamiliar or out-of-distribution actions based on the specific historical states, contributing to more targeted and context-aware regularization. In the states with high-certainty ($\left( u \approx 0 \right) \Rightarrow \alpha \approx \alpha_0$), the T-CQL will behave similar to the standard CQL with a mild penalty. However, in the rare or out‐of‐distribution states ($\left( u \gg 0 \right) \Rightarrow \alpha(s) \gg \alpha_0$), the T-CQL will strongly discourage the policy from selecting actions not seen in the dataset. During inference or offline evaluation, we can also inspect $u_\theta(s)$ as a warning flag so that more uncertain states will result in stronger conservatism. 

\subsubsection{Sequence Consistency Regularization}
We introduce step dropping from the original $L$-step state sequence to promote stable and robust Q-estimates in the presence of minor perturbations or noise in the historical input. This is similar to techniques frequently used in data augmentation or consistency losses for semi-supervised or unsupervised learning \citep{r32, r33}. we define the sequence consistency regularizer as follows: 
\begin{align}
  \Lcal_{\rm SC}
  = \E_{(s_{t-L+1:t},a,r,s')\sim\Dcal}
    \bigl\|Q_\theta(\Sbf_{\rm full}^{1:L},a)
         -Q_\theta(\Sbf_{\rm short}^{1:L-1},a)\bigr\|_2^2,
\end{align}
where $\Sbf_{\rm full}^{1:L}=(s_{t-L+1},\dots,s_t)$ and $\Sbf_{\rm short}^{1:L-1}=(s_{t-L+1},\dots,s_{t-1})$. Combining the above terms, the overall objective function can therefore be defined as:
\begin{align}
\label{formu:total_loss}
\mathcal{L}(\theta) = 
\underbrace{\mathcal{L}_{\text{TD}}}_{\text{Bellman error}} 
+ 
\underbrace{\Lcal_{\rm T-CQL}}_{\text{adaptive T-CQL regularizer}} 
+ 
\underbrace{\lambda_{\text{sc}} \, \mathcal{L}_{\text{SC}}}_{\text{consistency regularizer}}, 
\end{align}
where 
$\lambda_{\rm sc}\ge0$ controls sequence‐consistency strength, and target network parameters $\bar\theta$ are updated periodically via Polyak averaging. The procedure of T-CQL training can be seen in  Algorithm~\ref{alg:t-cql-training}.

\begin{algorithm}[t]
\caption{Training Procedure for Transformer‐based CQL (T‐CQL)}
\label{alg:t-cql-training}
\begin{algorithmic}[1]
\Require Offline dataset $\Dcal=\{(s_i,a_i,r_i,s'_i)\}_{i=1}^N$, sequence length $L$, discount $\gamma$, base CQL weight $\alpha_0$, uncertainty scale $\beta$, CQL temperature $\tau$, sequence consistency regularization weight $\lambda_{\rm sc}$, learning rate $\eta$, target update period $K$, total gradient steps $T$.
\State Initialize Q‐network parameters $\theta$ and target network $\bar\theta \leftarrow \theta$.
\For{$t = 1$ to $T$}
  \State Sample a minibatch $\{(s,a,r,s')\}\sim\Dcal$
  \Comment{batch size $B$ implied}
  \State Construct current‐state windows:
    \[
      \Sbf^{1:L} = \bigl(s_{t-L+1},\dots,s_t\bigr),\quad
      \Sbf'{}^{1:L} = \bigl(s_{t-L+2},\dots,s_{t+1}\bigr)
    \]
  \State \textbf{Forward pass:}
  \State \quad $(H,\,z,\,\bar z)\leftarrow \mathrm{TransformerEncoder}(\Sbf^{1:L})$
  \State \quad $Q\leftarrow\mathrm{MLP}([z\|\bar z])$  \Comment{$B\times|\Acal|$}
  \State \quad $u\leftarrow\mathrm{Var}\bigl(\psi_\theta(H)\bigr)$  \Comment{epistemic uncertainty}
  \State \quad Compute $Q'$ on $\Sbf'{}^{1:L}$ similarly
  \State \textbf{Compute losses in Equation (\ref{formu:total_loss})}
  \State \textbf{Backward \& update:}
  \State \quad $\theta \leftarrow \theta - \eta\,\nabla_\theta \Lcal$
  \If{$t \bmod K = 0$}
    \State $\bar\theta \leftarrow \tau\,\theta + (1-\tau)\,\bar\theta$
  \EndIf
\EndFor
\end{algorithmic}
\end{algorithm}

\subsection{Reward Design}
Here, a per timestep reward \(r_t\) was designed to capture both terminal outcomes and intermediate clinical improvements. Let \(s_t\) be the current patient's state and \(s_{t+1}\) is the next state. The total reward function can be used to minimize the indicators of VILI, and maximize their 90 day survival:
\[
r_t = 
\begin{cases}
+1, & \text{terminal } \& \text{ patient survives},\\
-1, & \text{terminal } \& \text{ patient dies},\\
R_{\rm inter}(s_t, s_{t+1}), & \text{otherwise},
\end{cases}
\]
where the terminal reward penalizes death at the final state of an episode with the value of -1 if the patients died in 90 days or +1 otherwise. Due to the sparse terminal reward, we further developed intermediate rewards
for non‐terminal transitions, which combines normalized improvements in APACHE‐II score \citep{r33} and driving pressure \citep{r7} during MV:
\[
\Delta_{\text{APACHE-II}} = \frac{\text{APACHE-II}_t - \text{APACHE-II}_{t+1}}{\text{APACHE-II}_{max}}, \]
\[
\Delta_{\text{DP}} = \frac{\text{DP}_t - \text{DP}_{t+1}}{\text{DP}_{max}}
\]
where $\text{APACHE-II}_{max}$ and $\text{DP}_{max}$ denotes the maximum values of APACHE-II and DP, respectively, found in the dataset. The positive values of $\Delta_{\text{APACHE-II}}$ and $\Delta_{\text{DP}}$ over time correspond to improved patient condition and reduced risk of VILI, respectively. The objective intermediate reward is then defined as:  
\[
R_{\rm inter}(s_t, s_{t+1}) = 
w_{\text{APACHE-II}} \Delta_{\text{APACHE-II}}
+
w_{\text{DP}} \Delta_{\text{DP}},
\]
where $w_{\text{APACHE-II}}$ and $w_{\text{DP}}$ are both a weighting factor, set to 0.5 in the study.  

\subsection{Dataset Description}
In this study, we used two publicly available datasets (MIMIC-III \citep{r36} and MIMIC-IV \citep{r37}) for offline RL model development and policy evaluation. Data were extracted using PostgreSQL (Version 16.0) and organized into two-hour time windows, including only adult patients (aged $\ge$ 18 years) who received at least two consecutive hours of pressure-controlled MV. Missing values were first addressed using forward-fill imputation, and any remaining were subsequently handled through k-nearest-neighbor (KNN) imputation. To avoid duplication from overlapping records in MIMIC-III and MIMIC-IV, we merged MIMIV‑III with only MIMIC‑IV patient records since 2013. The resulting state vectors contains 11,585 patients and 994,080 hours of PCV. Overall dataset was split into 80\% training and 20\% testing.

\subsection{Evaluation Methods}
For offline RL evaluation, we employed two approaches to estimate the performance of policies made by RL agents: 1) Using the conventional off-policy Fitted Q-Evaluation \citep{r34} based on offline static dataset, and 2) interactive validation based on the developed digital twins, by evaluating their response to the algorithm-suggested ventilator settings, representing a realistic ``at the bedside" closed-loop ventilation experience. 
\subsubsection{Fitted Q-Evaluation} 
The target of FQE is to fit the regression $y = r + \gamma \, Q_{\hat{\phi}}^{\pi}(s', \pi(s'))$ using a fixed learned policy $\pi$ for each transition \((s, a, r, s')\) in the offline dataset. The target netwrok $Q_{{\phi}}^{\pi}$ is obtained by minimizing $\mathbb{E}_{(s, a, r, s') \sim \mathcal{D}_{\text{train}}} \left[ 
\left( Q_{\phi}^{\pi}(s, a) - \left( r + \gamma Q_{\hat{\phi}}^{\pi}(s', \pi(s')) \right) \right)^2 
\right].$ Once $Q_{\phi}^{\pi}$ is fitted, the expected return of 
${\pi}$ is obtained via $\frac{1}{|\mathcal{D}_{\text{test}}|} \sum_{s \in \mathcal{D}_{\text{test}}} Q_{\phi}^{\pi}(s, \pi(s))$.
\subsubsection{Interactive Digital Twins}
To complement our off‐policy FQE estimation, we also deployed the fixed policy ${\pi}$ in 98 digital twins of ARF patients (see Figure~\ref{fig:pulmonary_model}) and measured their physiological responses to the suggested ventilator settings. For each simulated patient, we initialize the digital twin to their initial state  $s_0$ and simulate an episode by iteratively applying the policy $\pi$. At each time step, the next state is determined by the digital twin’s physiological responses to the set of actions selected by a greedy deterministic policy $\pi$ for the current state ($a = \arg\max_{a'} \pi(a' \mid s_t)$). To evaluate the policy, we used the cumulative reward over each simulated episode, and compared the levels of driving pressure only when clinically safe gas exchange targets were achieved, i.e. $\mathrm{PaO}_2 > 60\,\text{mmHg}$ and $\mathrm{PaCO}_2 < 60\,\text{mmHg}$.

\section{Results}
In this study we assessed the performance of the developed algorithm using both an off-policy FQE method and  interactive digital twins. The proposed T-CQL was compared to clinicians, and bench-marked against several popular offline RL methods such as Deepvent (CQL) \citep{r10}, Double Deep Q-Network (DDQN), and Batch-Constrained Deep Q-Learning (BCQ) \citep{r35}. 

\subsection{Off-policy Evaluation}

\subsubsection{Out-of-Distribution Performance}
\begin{figure}[htbp]
  \centering
  \includegraphics[width=0.47\textwidth]{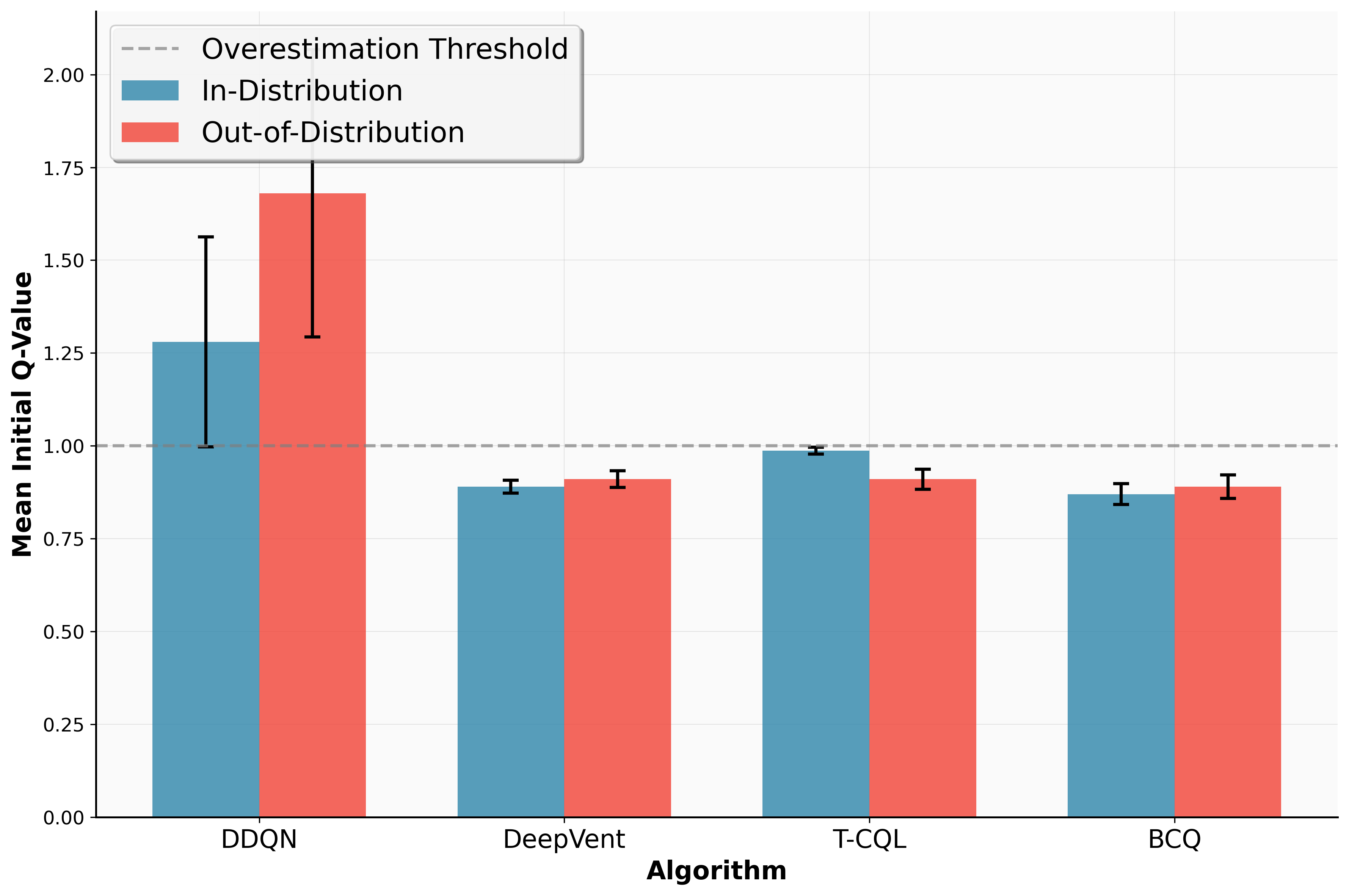}
  \caption{Comparison of average initial Q-values for in-distribution (ID) versus out-of-distribution (OOD) cases. Error bars denote the variance, and the horizontal line marks the maximum achievable expected return.}
  \label{fig:ood_compare}
\end{figure}

To evaluate the robustness of different methods in Out-of-Distribution (OOD) scenarios which may lead to dangerous ventilator settings in intensive care, we calculate the mean initial Q values for various RL policies, as shown in Figure \ref{fig:ood_compare}. Each RL policy on OOD scenarios are compared against the theoretical maximum return of \(1\) (in the absence of intermediate rewards).  Because FQE was trained on the same static dataset whose maximum episode return is \(1\), any \(\bar Q_0>1\) may introduce \emph{overestimation} under a distribution shift. DDQN showed significant overestimation in OOD scenarios, which is likely to generate overly optimistic and potentially unsafe clinical decisions. This is because DDQN is the only method presented that lacks an explicit mechanism to constrain bootstrapping errors or to ensure adherence to the clinician’s data distribution. In contrast, support‐constrained offline RL, e.g. Deepvent, BCQ, and T-CQL always remained below the threshold, as these methods enforce some form of “stay close to the data” or “penalize unfamiliar actions” constraint. Consequently, we excluded DDQN from further analyses.

\subsubsection{Quantitative Results}
As shown in Table 1, our proposed T-CQL model obtained the highest FQE score ($0.87 \pm 0.05$), significantly higher than the clinician policy ($0.46 \pm 0.01$) and conventional offline RL models (BCQ: $0.63 \pm 0.09$, Deepvent: $0.79 \pm 0.13$). Also, T-CQL yields the lowest estimated mortality score with lower variance ($0.16 \pm 0.02$). Moreover, as shown in Table 2, T-CQL demonstrates a stronger negative correlation ($-0.49 \pm 0.04$) between Q values and mortality compared to Deepvent and BCQ, indicating its superior performance in associating higher Q values with lower mortality rates and predicting patient outcomes.

\begin{table}[ht]
\centering
\begin{tabular}{lcc}
\toprule
\textbf{Model} & \textbf{FQE Score} & \textbf{Mortality Score}\\
\midrule
Clinician         & $0.46 \pm 0.01$   & $0.27 \pm 0.05$  \\
BCQ          & $0.63 \pm 0.09$   & $0.24 \pm 0.05$ \\
Deepvent (CQL)    & $0.79 \pm 0.13$   & $0.19 \pm 0.03$ \\
T-CQL(This study)                     & $0.87 \pm 0.05$  & $0.16 \pm 0.02$ \\
\bottomrule
\end{tabular}
\caption{Comparison of expected return and estimated mortality score for various policies. The values are presented as means $\pm$ standard deviations across five runs. Higher FQE scores reflect greater expected cumulative return under the policy (i.e. better long‐term patient outcome trajectories), whereas lower estimated mortality rates correspond to higher predicted survival likelihood}
\label{tab:expected_return_mortality}
\end{table}

\begin{table}[ht]
\centering
\begin{tabular}{lc}
\toprule
\textbf{Model}& \textbf{Pearson Correlation Coefficient}\\
\midrule
BCQ           & $-0.30 \pm 0.07$\\
Deepvent (CQL)     & $-0.38 \pm 0.03$\\
T-CQL(This study)                      & $-0.49 \pm 0.04$\\
\bottomrule
\end{tabular}
\caption{ The Pearson correlation coefficients  between the expected return and estimated mortality (p$<$0.0001).}
\label{tab:expected_return_mortality}
\end{table}

\subsubsection{Action Distributions of RL Policies vs. Clinicians on Offline Dataset} 
\Cref{fig:figA1_action_distribution} shows the action‐distribution for each policy. The T-CQL agent most closely mirrors clinician practice, preferring lung-protective settings, where lower PEEP (5-10~cmH$_2$O) and moderate Pvent (15–25~cmH$_2$O) are usually preferred. Other favored settings include FiO$_2$ of 45–55\%, respiratory rates of 20–22 breaths/min, and I:E ratios of 1:3 to 1:2. Across all offline variants of RL, T-CQL best reproduces the distribution of ventilator settings prescribed by clinicians.

\begin{figure*}[!htbp]
  \centering
  \includegraphics[width=1.02\textwidth]{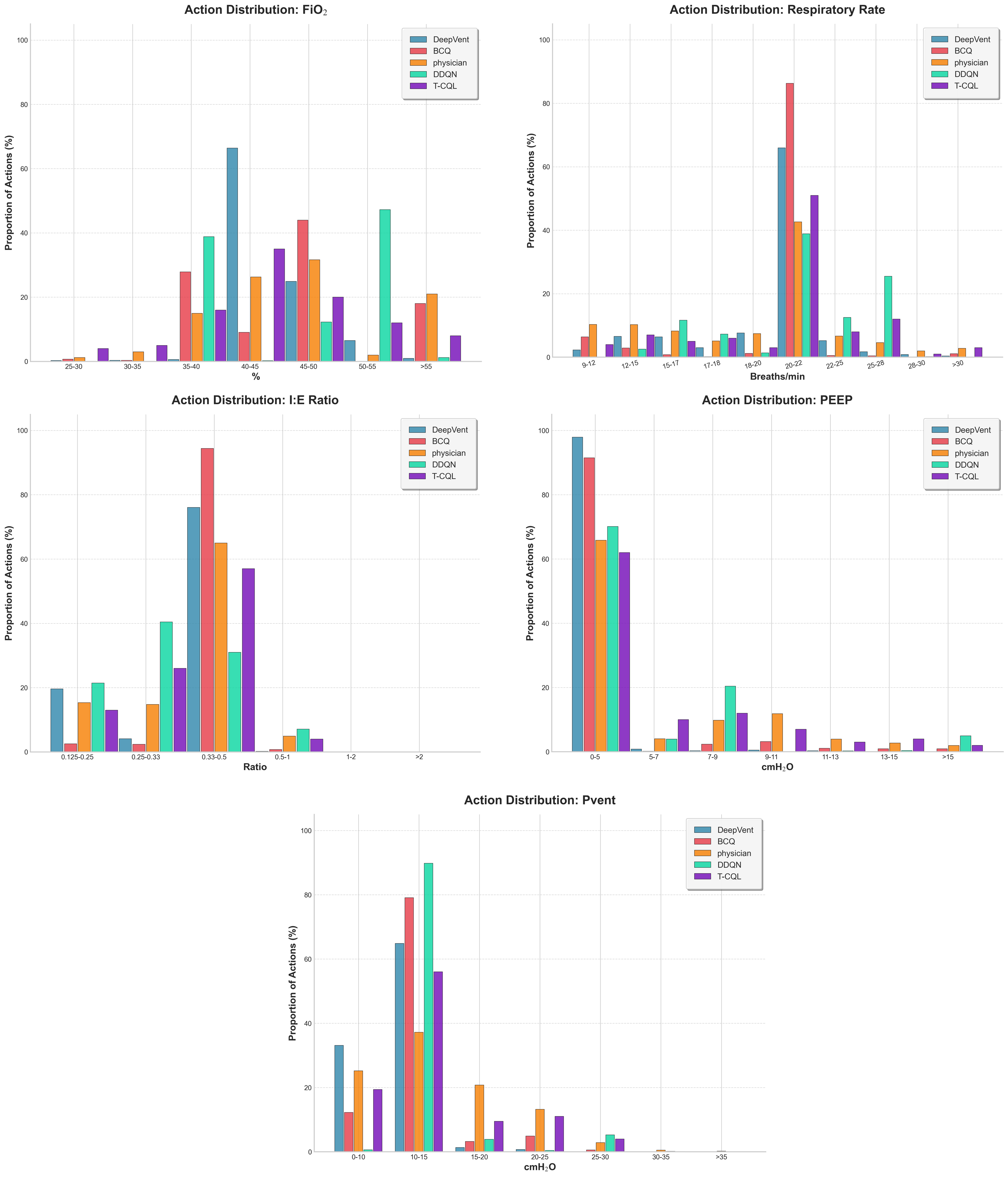}
  \caption{Distribution of ventilator settings chosen by different policies on the MIMIC test set. "Clinicians" refer to the actions taken by clinicians in the original MIMIC dataset.}
  \label{fig:figA1_action_distribution}
\end{figure*}
\subsubsection{Action Distributions of RL Policies vs. Clinicians on
Digital Twins}
Figure ~\ref{fig:figA1_action_distribution} shows the marginal distributions of five ventilator parameters—FiO$_2$, respiratory rate (RR), I:E ratio, PEEP and inspiratory pressure (Pvent)—for four RL agents ((DeepVent, BCQ, DDQN, T-CQL)), the original MIMIC‐III clinician policy, and an external cohort of RWTH intensivists' policy. Each bar gives the percentage of 2-hour epochs in which that setting‐range was chosen.
T-CQL consistently reproduces the nuanced ventilator‐setting distributions of experienced clinicians while modestly expanding the safe range (e.g., slightly higher PEEP). RWTH intensivists often set 9–11 cm H$_2$O or 11–13 cm H$_2$O (~28–29 \% each), reflecting a more aggressive approach. These results highlight T-CQL’s ability to learn balanced, lung-protective strategies from offline data, suggesting strong potential as a clinician-in-the-loop decision support tool to standardize best practice in invasive mechanical ventilation.

\subsection{Online Evaluation Using Interactive Digital Twins}
\subsubsection{Compliance with Targets on Gas Exchange and Lung-Protective Ventilation}
The goal of the online evaluation is to achieve the following criteria to ensure both adequate gas change and lung protective strategies based on expert clinical guidance:
\begin{itemize}
    \item \textbf{Gas Exchange Constraint}: Ensure $\mathrm{PaO}_2 \geq 60$ mmHg and $\mathrm{PaCO}_2 \leq 60$ mmHg to prevent hypoxemia and hypercapnia.
    \item \textbf{Peak Pressure Constraint}: Ensure PIP $\leq 35$ cmH$_2$O to avoid barotrauma.
\end{itemize}
\begin{table}[ht]
\centering
\begin{tabular}{lcc}
\toprule
\textbf{Model} & \shortstack{\textbf{Safety Constraints} \\ \textbf{Rate (\%)}} & \shortstack{\textbf{Reduced DP} \\ \textbf{Rate (\%)}} \\
\midrule
Clinician  & $43.88 \pm 0.15$  & $37.76 \pm 0.02$\\
DDQN              & $22.45 \pm 2.58$  & $20.41 \pm 1.36$\\
BCQ           & $41.83 \pm 1.47$ & $40.82 \pm 0.95$\\
Deepvent (CQL)     & $34.69 \pm 1.63$ & $33.67 \pm 1.05$\\
T-CQL(This study)    & $47.96 \pm 1.15$ & $44.90 \pm 0.92$\\
\bottomrule
\end{tabular}
\caption{Compliance rates for safety constraints, and reduction in driving pressure (DP), for each policy. Values are presented as means ± standard deviations across five runs. Safety constraints compliance rate indicates the proportion of times all safety constraints were met. The Reduced DP rate indicates the proportion of times the driving pressure was reduced compared to the baseline value (initial state).}
\label{tab:expected_return_mortality}
\end{table}
 As shown in Table 3, the T‐CQL agent achieved the highest compliance rate at $47.96 \pm 1.15$, outperforming the clinician policy from the MIMIC dataset ($43.88 \pm 0.15$), and offline RL policies, e.g., BCQ ($41.83 \pm 1.47$), and DeepVent ($34.69 \pm 1.63$). Interestingly, this level of compliance does not align with the FQE scores, where offline RL models achieved significantly higher values than the clinician policy but demonstrated much lower compliance rates when applied in digital twins of real patients. This discrepancy highlights a key limitation of the current OPE methods in reliably assessing model performance in dynamic treatment settings. In contrast, T-CQL is the only method that achieved both a high FQE score and a high compliance rate in simulated real-world scenarios. This suggests that T‐CQL more consistently achieves both goals of maintaining adequate gas exchange while applying a lung-protective ventilation strategy.

To assess the T-CQL model performance in providing a lung-protective strategy that helps reduce the risk of VILI, we evaluated the proportion of times DP was reduced compared to the initial state, i.e. where the suggested strategy led to a lower DP compared to the patient's baseline. We found that nearly all patients who met the oxygenation targets under the T-CQL policy also experienced a reduction in DP. These findings suggest that, for most patients who met gas exchange targets, T-CQL’s recommended ventilator settings were also at least as protective (and in many instances slightly more optimized) as those applied by clinicians.
\subsubsection{Action Distributions of RL Policies vs. Clinicians on Digital Twins}
S\Cref{fig:figA1_action_distribution} compares the marginal distributions of five ventilator settings, e.g. PEEP, FiO$_2$, RR, I:E ratio, and Pvent, across four offline RL policies, alongside the clinician strategies from the both MIMIC dataset and the University Hospital Aachen. Overall, T-CQL most closely replicates the distribution of ventilator settings chosen by MIMIC clinicians, favoring moderate levels of PEEP, FiO$_2$, RR, and Pvent while avoiding the extreme values observed in less constrained RL agents such as DDQN. This indicates that, among all offline RL models, T-CQL most effectively learns meaningful strategies from the offline dataset, achieving a better balance between maintaining adequate gas exchange and minimizing the risk of VILI.
\begin{figure*}[!htbp]
  \centering
  \includegraphics[width=1.02\textwidth]{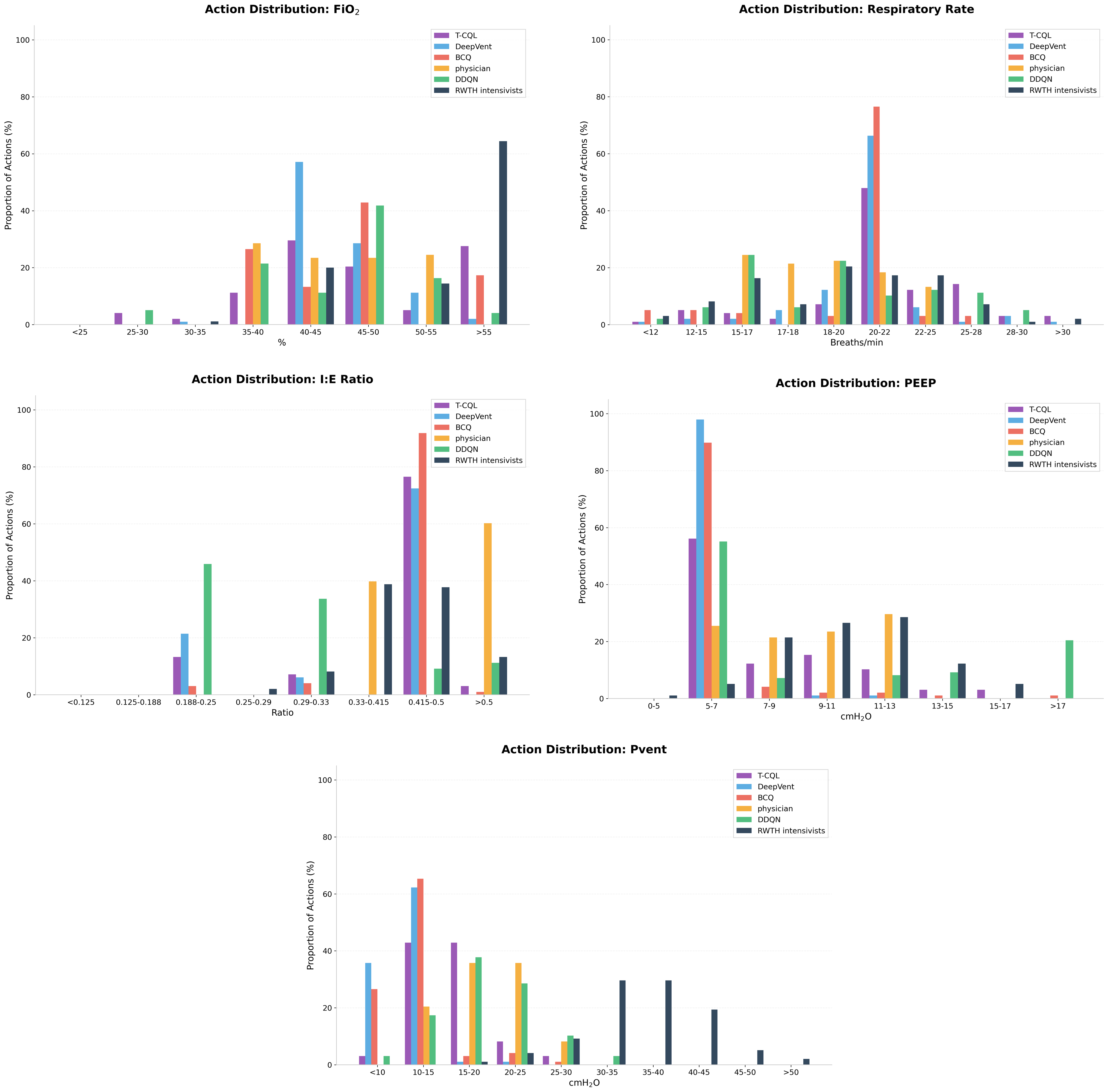}
  \caption{Distribution of ventilator settings selected by different policies across digital twins of real patients. "Physician" refers to clinician-derived actions from the MIMIC dataset, while "RWTH Intensivists" represent ventilator settings applied by clinicians at the University Hospital of Aachen.}
  \label{fig:figA1_action_distribution}
\end{figure*}

\section{Discussion and Conclusion}
The T-CQL algorithm proposed in this study outperformed other existing offline state-of-the-art RL models in the application of automated and personalized MV, by achieving higher expected returns, lower predicted mortality, robust OOD behavior, and superior compliance with gas exchange and lung-protective ventilation constraints. The model performance was evaluated using both offline FQE and online interactive evaluation with digital twins of real patients. Across both OPE and digital twins, T-CQL closely followed clinician practice. These results demonstrate that a conservative, temporally aware RL approach can generate ventilator settings that optimize gas exchange while minimizing the risk of VILI. Also, the discrepancy in performance between OPE metrics and online outcomes obtained from evaluation via digital twins highlights the limitations of retrospective evaluation under domain shift, policy divergence, and the inherent trade-off between protective ventilation and adequate oxygenation. 

\bibliography{aaai2026}    

\end{document}